\documentclass[runningheads]{llncs}

\usepackage{graphicx}
\usepackage{booktabs}
\usepackage{amsmath,amssymb}
\usepackage{algorithm}
\usepackage{algpseudocode}
\usepackage{float}
\usepackage[dvipsnames]{xcolor}
\usepackage[breaklinks,colorlinks,hypertexnames=false,%
            linkcolor=blue!50!black,%
            citecolor=blue!50!black,urlcolor=blue!50!black]{hyperref}

\providecommand{\citep}[1]{\cite{#1}}
\providecommand{\citet}[1]{\cite{#1}}

\newcommand{\methodname}{\textsc{FLITE}}
\newcommand{\methodlong}{Federated Low-rank Iterative Training Engine}

\graphicspath{{figs/}}
\DeclareGraphicsExtensions{.pdf,.PDF}

\begin{document}

\title{Federated Lightweight Fine-Tuning}
\titlerunning{Federated Lightweight Fine-Tuning}

\author{Radhakrishna Achanta \and Will Reed}
\authorrunning{R. Achanta and W. Reed}
\institute{Cisco Systems Inc., USA\\
  \email{\{rachanta,wilreed\}@cisco.com}}

\maketitle

\begin{abstract}
Federated fine-tuning is bottlenecked by communication: FedAvg and pseudo-gradient
schemes transmit a payload that scales with the model, and gradient compression
shrinks it by only a constant factor. We take a different lever. Mapping networks
generate a network's weights from a small trainable latent through a frozen affine
projection; because the map is shared and affine, averaging latents is
\emph{exactly} averaging the generated weights. We turn this into a practical
low-bandwidth federated channel with two changes: a \emph{low-rank,
seed-regenerable} factorisation of the projection (cutting generator memory from
$\sim$$80$\,GB to $\sim$$10$\,MB), and a \emph{delta} formulation
$\theta = \theta^{\mathrm{pre}} + UV^{\top}z$ that learns an additive correction
around a shared centrally-pretrained base --- federated fine-tuning, which is
what makes the method work at scale. A frozen orthogonal classifier head further
removes the head from the payload while improving accuracy. On CIFAR-100 with
ResNet-18+GroupNorm, our method (\methodname{}, \methodlong{}) communicates
$1{,}280$ floats ($\approx 5$\,KB) per client per round --- an $8718\times$
reduction --- and reaches $74.67\%$, within $\approx 0.5$\,pp of full-weight
FedAvg. The averaging identity holds to floating-point precision
($6{\times}10^{-8}$); the method sits one to two orders of magnitude below
PowerSGD and top-$k$ on the bandwidth--accuracy Pareto; it matches or exceeds
full-weight FedAvg under strong non-IID skew. int4 latents reach $648$ bytes per
round at unchanged accuracy, whereas int4 full-weight FedAvg collapses to chance.
\end{abstract}

\section{Introduction}
\label{sec:intro}

Federated and distributed training let many workers improve a shared model
without pooling their data \citep{mcmahan2017communication,kairouz2019advances},
but they pay a steep communication cost. The dominant paradigm exchanges
model state directly: FedAvg \citep{mcmahan2017communication} transmits the model
itself, and pseudo-gradient schemes such as DiLoCo \citep{douillard2023diloco}
transmit a full-size weight delta acting as an outer-loop update. Both have
per-round bandwidth $\Theta(|W|)$, the number of model parameters, so for a modern
network the message --- tens of megabytes for even a small ResNet --- dominates the
cost of federation.

A structural assumption underlies these methods: bandwidth is reduced along the
\emph{frequency} axis --- communicating less often --- while the size of each
message is treated as fixed. Even dedicated gradient-compression methods
(low-rank projection \citep{vogels2019powersgd}, top-$k$ sparsification
\citep{aji2017sparse}, quantization \citep{alistarh2017qsgd}) only shrink the
fixed message by a constant factor, still transmitting an object of size
proportional to $|W|$, and lowering synchronisation frequency trades amortised cost
for client drift, which is most damaging precisely when data are heterogeneous.
This paper takes the orthogonal \emph{size} axis: we keep communication frequent
but make each message a tiny, model-independent latent, while preserving the exact
FedAvg averaging semantics in weight space.

\paragraph{Mapping networks: a different lever.} A recent and very different idea,
\emph{mapping networks}, trains a small \emph{latent} vector $z$ that, through a
fixed (frozen) projection, generates the weights of a much larger network. The
latent is a compact carrier of model state. Crucially, when the generator is
affine and shared, \emph{averaging latents is exactly averaging the generated
weights} --- so a federated client could communicate only the small latent while
the server aggregation remains ordinary FedAvg. The idea is attractive but, as
proposed, has two blockers: (i) the projection matrix is itself huge (on the order
of tens of gigabytes for a ResNet-18), and (ii) the latent cannot drive a large
network when trained from scratch (it collapses to near-chance accuracy).

\paragraph{Our approach.} \methodname{} (\methodlong{}) turns the mapping-network latent into a practical
low-bandwidth communication channel by addressing both blockers and re-targeting
the method to where it is strong. First, we replace the dense projection with a
\emph{low-rank, seed-regenerable} factorisation, cutting generator memory from
$\approx 80$\,GB to $\approx 2$\,GB (or $\approx 10$\,MB if regenerated from a seed)
and the regeneration compute correspondingly. Second, rather than asking the latent
to encode a whole network, we use it to encode an \emph{additive delta around a
shared, centrally-pretrained frozen base} --- federated \emph{fine-tuning} rather
than from-scratch training. This keeps the exact averaging identity, sidesteps the
from-scratch failure, and exploits the fact that the correction a well-trained model
needs is empirically low-dimensional. Finally, we freeze an \emph{orthogonally
initialised} classifier head, which removes it from the communicated payload and,
as an ablation shows, improves accuracy.

\paragraph{Results.} On CIFAR-100 with ResNet-18+GroupNorm, our latent communicates
$1{,}280$ floats ($\approx 5$\,KB) per client per round --- an $8718\times$ reduction
--- and reaches $74.67\%$, within $\approx 0.5$\,pp of full-weight FedAvg and
statistically at the centralized baseline. The dimension traces a controllable
bandwidth--accuracy frontier; the method sits one to two orders of magnitude below
PowerSGD and top-$k$ on the Pareto; it matches or exceeds full-weight FedAvg under
strong non-IID skew and is stable across client counts. The frozen orthogonal head
adds $+0.54$\,pp at zero communication. Being low-dimensional, the latent is highly
quantization-robust: int4 latents reach $648$ bytes per round at unchanged accuracy,
while int4 full-weight FedAvg collapses to chance.

\paragraph{Contributions.}
\begin{itemize}
  \item A \textbf{delta low-rank mapping} $\theta = \theta^{\mathrm{pre}} + UV^{\top}z$
        that preserves the \emph{exact} FedAvg averaging identity (verified to
        $6{\times}10^{-8}$) while reducing generator memory by $\sim$$40\times$
        (further to $\sim$$10$\,MB via seed regeneration).
  \item A \textbf{low-bandwidth federated fine-tuning protocol} that transmits a
        $\sim$$5$\,KB latent per round at parity accuracy, with a favourable
        bandwidth--accuracy Pareto against PowerSGD and top-$k$, and improved
        robustness under heterogeneity and small client counts.
  \item The \textbf{frozen orthogonal classifier} as a free, composable design
        choice that removes the head from the payload and improves accuracy;
        \textbf{int4 quantization-robustness} of the latent channel at $648$
        bytes per round, where matched-bit full-weight FedAvg collapses.
  \item \textbf{Negative results} showing that mapping networks as a
        stand-alone training method collapse at ResNet-18 scale regardless of
        dimension, rank, or warm start, identifying federated fine-tuning as
        the regime in which the idea does pay off (\S\ref{ssec:res-scratch}).
\end{itemize}

The remainder of the paper presents the method (\S\ref{sec:method}), the setup
(\S\ref{sec:setup}), results (\S\ref{sec:results}), and concluding remarks on
scope and limitations (\S\ref{sec:discussion}).

\section{Related Work}
\label{sec:related}

\paragraph{Federated averaging and its variants.} FedAvg
\citep{mcmahan2017communication} established the dominant paradigm: clients train
locally and the server averages their models. Subsequent work improves robustness
to heterogeneity and client drift --- FedProx \citep{li2018federated} adds a
proximal term, SCAFFOLD \citep{karimireddy2019scaffold} uses control variates,
Matched Averaging \citep{wang2020federated} aligns neurons before averaging, and
adaptive server optimisation \citep{reddi2020adaptive} stabilises aggregation. All
of these communicate an object of size $\Theta(|W|)$ each round; they change
\emph{how} weights are aggregated, not the \emph{size} of the message. Our method
is complementary: it preserves the FedAvg averaging rule exactly but shrinks the
message to a latent. We further note a setting distinction: FedAvg, FedProx, and
DiLoCo are typically run as \emph{from-scratch} federated training, with the
server initialising $\theta_0$ and the federation itself producing the trained
model. Our scheme operates in the \emph{federated fine-tuning} regime, in which
every client --- ours and the FedAvg baseline alike --- starts from the same
shared pretrained checkpoint. All bandwidth and accuracy comparisons in
\S\ref{sec:results} use this matched setting, so the contrast is between two
fine-tuning protocols at different bandwidths, not between a fine-tuning method
and a from-scratch one.

\paragraph{Low-frequency and pseudo-gradient methods.} Local SGD
\citep{stich2018local,khaled2020tighter} and DiLoCo \citep{douillard2023diloco}
reduce communication by synchronising less often, performing many local steps
between merges. This lowers amortised cost but transmits a full-size update at each
merge and risks client drift, particularly under heterogeneity. Our latent is small
enough that frequent synchronisation is affordable, which we show converts into a
robustness advantage at a fixed byte budget (\S\ref{ssec:res-k}).

\paragraph{Communication compression.} A large literature compresses the
transmitted gradient/update: low-rank projection (PowerSGD \citep{vogels2019powersgd}),
magnitude sparsification (top-$k$ \citep{aji2017sparse}, deep gradient compression
\citep{lin2018deep}), and quantization (QSGD \citep{alistarh2017qsgd}, signSGD
\citep{bernstein2018signsgd}), often with error feedback to preserve convergence
\citep{karimireddy2019error}. These methods reduce the payload by a constant factor
but the compressed object is still derived from, and scales with, $|W|$. We compare
directly against PowerSGD and top-$k$ and show our latent is one to two orders of
magnitude cheaper at matched accuracy (\S\ref{ssec:res-pareto}); the two approaches
are moreover composable (a latent can itself be quantized).

\paragraph{Subspace training, low-rank adaptation, and mapping networks.} Training
in a low-dimensional random subspace is known to suffice for many objectives, an
observation formalised by intrinsic-dimension studies \citep{li2018intrinsic} and
exploited for parameter-efficient fine-tuning by LoRA \citep{hu2021lora}. Mapping
networks \citep{mappingnetworks2026} take this further, generating a network's
full weights from a low-dimensional latent through a frozen generator whose core
is an affine projection. We adopt the \emph{affine special case} of the
mapping-network generator and differ in three ways: (i)~we factor the projection
into a low-rank, seed-regenerable form (the original materialises a dense
projection), removing its prohibitive memory and compute; (ii)~we operate in
delta mode on a shared pretrained base, which is what makes the method work at
scale (from-scratch mapping fails, \S\ref{ssec:res-scratch}); and (iii)~we use it
as a federated \emph{communication} channel, exploiting the exact averaging
identity that the affine map admits. Unlike LoRA, our factors are frozen and
shared so that averaging latents equals averaging weights --- the property
federated aggregation requires.

\paragraph{Optimization-centric decentralized FL and function-space methods.}
A parallel line improves decentralized FL through optimisation, topology, and
personalisation while remaining in parameter space
\citep{dspodfl2025,dpfl2025,ntkdfl2025,fedspd2025,gflat2026}; their messages
still scale with $|W|$ and our latent channel is complementary.
A distinct line collaborates in \emph{function space}, exchanging predictions on
a shared probe set (FedMD~\citep{fedmd2019}, FedDF~\citep{lin2020ensemble}).
These decouple the message from $|W|$ and accommodate model heterogeneity, but
optimise a different (distillation) objective and are not directly comparable
to exact parameter-space averaging. Our channel is weight averaging, carried
out exactly in a low-dimensional latent.

\paragraph{Frozen classifiers and normalization.} Fixing the classifier head is
known to be largely harmless \citep{hoffer2018fix} and can be beneficial with
well-conditioned geometries such as regular-polytope/ETF
classifiers~\citep{pernici2021regular} motivated by neural
collapse~\citep{papyan2020prevalence}; FedBABU~\citep{oh2022fedbabu} freezes the
head during federation. We use a frozen orthogonal head as a composable
component that removes the classifier from the payload and improves accuracy
(\S\ref{ssec:res-ablations}). BatchNorm statistics are not naturally averageable
across heterogeneous clients, motivating GroupNorm~\citep{wu2018group} or local
BatchNorm~\citep{li2021fedbn} in federated settings~\citep{hsieh2020noniid}; we
adopt GroupNorm and accept absolute accuracy below BatchNorm SOTA.

\section{Method}
\label{sec:method}

We first recall the mapping-network generator and the property that makes it
attractive for federated averaging (\S\ref{ssec:recap}). We then identify the
two obstacles that prevent it from being used directly at the scale of a modern
network (\S\ref{ssec:obstacles}), and present our delta low-rank parameterisation
that removes them (\S\ref{ssec:delta}). Finally we describe an optional
seed-regenerated projection (\S\ref{ssec:seed}), the frozen orthogonal classifier
(\S\ref{ssec:fc}), and the full federated protocol (\S\ref{ssec:protocol}).

\subsection{Mapping networks and the averaging identity}
\label{ssec:recap}

Mapping networks~\citep{mappingnetworks2026} generate the weights of a target
layer from a small trainable \emph{latent} vector $z$ through a fixed (frozen)
generator. The original proposal augments this map with modulation and a
nonlinearity; we study its affine special case, which is the piece that admits
an exact averaging identity. For a layer with
$P$ parameters and $z \in \mathbb{R}^{d}$, $d \ll P$,
\begin{equation}
  \theta \;=\; b \;+\; W_m\, z,
  \label{eq:affine-generator}
\end{equation}
where $b \in \mathbb{R}^{P}$ is a fixed base vector and $W_m \in \mathbb{R}^{P\times d}$
is a fixed projection; only $z$ is trained. In the original formulation $W_m$ is a
\emph{dense} random matrix, materialised in full. Because the map
$z \mapsto \theta$ is affine and $b, W_m$ are \emph{shared} across all clients,
averaging latents is identical to averaging the generated weights:
\begin{equation}
  \frac{1}{K}\sum_{k=1}^{K}\bigl(b + W_m\, z_k\bigr)
  \;=\; b + W_m\!\left(\frac{1}{K}\sum_{k=1}^{K} z_k\right).
  \label{eq:identity}
\end{equation}
\emph{This is exactly the FedAvg update, carried out in the $d$-dimensional latent
space instead of the $P$-dimensional weight space.} A client therefore needs to
transmit only $z$ ($d$ floats) rather than $\theta$ ($P$ floats), and the server
aggregation is unchanged. Dropping the modulation term and any post-map
nonlinearity is thus a deliberate design choice: it preserves
Eq.~\eqref{eq:identity} by construction.

\subsection{Two obstacles at scale}
\label{ssec:obstacles}

\paragraph{Memory and compute.} In the original mapping network the projection
$W_m$ is a dense matrix of shape $P \times d$ \emph{per layer}. For a ResNet-18
($P \approx 11.2$M aggregated over layers) even a modest $d$ makes $W_m$ enormous:
materialising the dense projections requires on the order of $80$\,GB in
\texttt{fp32}, far exceeding a single accelerator, and
every weight regeneration is a dense $P \times d$ matrix--vector product. The
latent is small, but the dense generator that expands it is not.

\paragraph{From-scratch training.} When $b$ is a random initialisation and the
latent must encode the \emph{entire} network, optimisation collapses at
ResNet-18 scale: training the latent from a Kaiming-initialised base reaches only
$\approx 2.5\%$ test accuracy on CIFAR-100 (\S\ref{sec:results}). The latent does
not have the capacity to drive a large network from scratch.

\subsection{Delta low-rank mapping}
\label{ssec:delta}

We address both obstacles with a single change of parameterisation. Rather than
generating each layer's weights from scratch with a dense projection, we (i)
generate an additive \emph{delta} on top of a frozen, centrally-pretrained
backbone, and (ii) replace the dense $W_m$ with a \emph{low-rank} factorisation
$W_m = U V^{\top}$ of rank $r \ll d$. The low-rank factorisation is \emph{not} part
of the original mapping network --- which materialises $W_m$ densely --- but is
introduced here specifically to cut both the storage of the generator and, by
consequence, the cost of every weight-regeneration matrix--vector product
(\S\ref{ssec:obstacles}); it is also what makes the small-$d$ regime that drives
the per-round payload down practical. For each mapped layer $l$,
\begin{equation}
  \theta_l \;=\; \underbrace{\theta^{\mathrm{pre}}_l}_{\text{frozen base}}
  \;+\; U_l\, V_l^{\top}\, z_l,
  \qquad
  U_l \in \mathbb{R}^{P_l\times r},\;
  V_l \in \mathbb{R}^{d_l\times r},
  \label{eq:delta}
\end{equation}
where $\theta^{\mathrm{pre}}_l$ is the (flattened) pretrained weight of layer $l$,
$U_l, V_l$ are \emph{frozen} factors with orthonormal columns, and only the
per-layer latent $z_l \in \mathbb{R}^{d_l}$ is trained. The effective projection is
$W_m^{(l)} = U_l V_l^{\top}$ of rank at most $r$. We initialise the latent at zero
($z_l = 0$), so the network at the start of mapping is \emph{exactly} the
pretrained model; the latent only learns refinements.

\paragraph{The averaging identity still holds exactly.} Equation~\eqref{eq:delta}
is affine in $z_l$ with a shared frozen offset $\theta^{\mathrm{pre}}_l$ and shared
frozen factors $U_l, V_l$. The same algebra as Eq.~\eqref{eq:identity} applies
per layer, and the pretrained offset cancels in the mean:
\begin{equation}
  \frac{1}{K}\sum_k \bigl(\theta^{\mathrm{pre}}_l + U_l V_l^{\top} z_{l,k}\bigr)
  = \theta^{\mathrm{pre}}_l + U_l V_l^{\top}\!\Bigl(\tfrac{1}{K}\textstyle\sum_k z_{l,k}\Bigr).
  \label{eq:delta-identity}
\end{equation}
We verify this numerically end-to-end: the maximum discrepancy between averaging
latents and averaging the generated weights is $5.96\times 10^{-8}$, i.e.\
floating-point round-off.

\paragraph{Memory and compute.} The factored projection costs $r\,(P_l + d_l)$
parameters per layer instead of $P_l\,d_l$, and the regeneration becomes two small
matrix--vector products ($V_l^{\top} z_l$ then $U_l(\cdot)$) costing
$\mathcal{O}\bigl(r(P_l + d_l)\bigr)$ rather than $\mathcal{O}(P_l d_l)$ flops.
Aggregated over ResNet-18 at $r=32$ the storage is
$\approx 2$\,GB rather than $\approx 80$\,GB --- a reduction of more than an order
of magnitude that brings the generator onto a single device. Crucially, the low rank does \emph{not} cost accuracy in
delta mode: varying $r \in \{8, 32, 128, 256\}$ changes test accuracy by
$<0.2$\,pp (\S\ref{sec:results}), because the delta a well-pretrained model needs
is itself low-dimensional.

\paragraph{Why delta mode resolves the from-scratch failure.} The latent no longer
has to represent the whole network, only the small correction around a strong
base. This both makes optimisation tractable and explains why a tiny latent
suffices: with $d_l = 64$, $r = 32$ the entire ResNet-18 is steered by
$1{,}280$ trainable floats (20 mapped layers $\times\,64$), an $8718\times$
reduction relative to the $11.2$M parameters, while recovering the pretrained
accuracy (\S\ref{sec:results}).

\subsection{Seed-regenerated projection (optional)}
\label{ssec:seed}

The factors $U_l, V_l$ are deterministic functions of a shared random seed
(Gaussian draws orthonormalised by QR): a single $64$-bit seed regenerates
them identically on every client, preserving Eq.~\eqref{eq:delta-identity}
exactly. This trades recompute for a further memory reduction, from $\approx
2$\,GB of stored factors to $\approx 10$\,MB of peak working memory.

\subsection{Frozen orthogonal classifier}
\label{ssec:fc}

We initialise the final classifier (the fully-connected head) with an orthogonal
weight matrix and \emph{freeze} it. This removes the classifier from the set of
mapped/communicated parameters entirely, and --- as an ablation confirms
(\S\ref{sec:results}) --- is accuracy-positive rather than merely neutral: on
CIFAR-100 the frozen orthogonal head reaches $73.53\%$ versus $72.99\%$ for a
trainable head and $72.95\%$ for a frozen Kaiming head. An orthonormal, fixed set
of class prototypes gives a well-conditioned target geometry for the backbone to
align to, while contributing zero communication.

\subsection{Federated protocol}
\label{ssec:protocol}

The complete scheme is a drop-in replacement for FedAvg in which the per-round
payload is the latent rather than the model. The server broadcasts the frozen
pretrained base $\theta^{\mathrm{pre}}$ and the seed \emph{once}; thereafter each
round transmits only $z$ (Algorithm~\ref{alg:latentfed}). With $K$ clients and
$D = \sum_l d_l$ latents, one round transfers $2 K D \cdot 4$ bytes: $\approx
5$\,KB per client per direction for ResNet-18 with $D = 1{,}280$, against
$\approx 45$\,MB for the full $11.2$M-parameter model.

\begin{algorithm}[t]
\caption{\methodname{} federated round with optional latent quantization. All
clients cache $\theta^{\mathrm{pre}}$ and factors $\{U_l, V_l\}$ (or the shared
seed that regenerates them, \S\ref{ssec:seed}); factors live on the mapped
weight tensors only. Non-mapped parameters (FC head, biases, normalization)
remain frozen. The generator $\theta_l(z_k) = \theta^{\mathrm{pre}}_l + U_l V_l^{\top} z_{k,l}$
is re-evaluated on every forward/backward pass; initialise $\bar z \gets 0$.
Aggregation is the unweighted mean; sample-weighted $p_k \propto n_k$ preserves
the identity too.}
\label{alg:latentfed}
\begin{algorithmic}[1]
\State \textbf{Server} broadcasts $\bar z$ to clients $k \in \mathcal{S}$.
\For{each client $k \in \mathcal{S}$ \textbf{in parallel}}
  \State Set $z_k \gets \bar z$.
  \State Train $z_k$ for $E$ local epochs on client data $\mathcal{D}_k$.
  \State \emph{(Optional)} Quantize $z_k$ to $q$-bit integers (uniform-affine, per-tensor scale).
  \State Upload $z_k$ ($4D$ bytes at fp32, $qD/8$ at $q$-bit; $D$ is the total latent size).
\EndFor
\State \textbf{Server} takes the unweighted mean of the (dequantized) uploads.
\end{algorithmic}
\end{algorithm}

\section{Experimental Setup}
\label{sec:setup}

\paragraph{Models and datasets.} Our primary setting is \textbf{ResNet-18 with
GroupNorm} ($\approx 11.2$M parameters) on \textbf{CIFAR-100}. We additionally
evaluate ResNet-18+GN on \textbf{CIFAR-10} and, to test scaling in model and label
space, \textbf{ResNet-34 with GroupNorm} ($\approx 21.3$M parameters) on
\textbf{TinyImageNet} (200 classes). To probe a non-vision modality we additionally
run a compact \textbf{TinyGPT} ($6$ layers, $384$-dim embeddings, $\approx 11.5$M
trainable parameters) on \textbf{WikiText-2} language modelling. We use GroupNorm rather than BatchNorm
throughout because BatchNorm running statistics are not naturally averageable
across clients in federated learning, whereas GroupNorm carries no such buffers
\citep{wu2018group,hsieh2020noniid}. We note up front that GroupNorm places the
\emph{absolute} accuracies somewhat below BatchNorm SOTA; our claims concern
\emph{relative} parity at greatly reduced bandwidth, not absolute state of the art.

\paragraph{Centrally-pretrained base.} All methods share a single centrally
pretrained backbone, trained with SGD (momentum $0.9$, learning rate $0.05$ with a
$3$-epoch warmup and cosine decay, weight decay $5\times10^{-4}$, batch size $128$)
for $100$ epochs, with the orthogonal classifier head frozen from initialisation
(\S\ref{ssec:fc}). The base reaches $94.4\%$ on CIFAR-10, $74.6\%$ on CIFAR-100,
and $62.5\%$ on TinyImageNet. The TinyGPT base reaches validation perplexity
$66.1$ on WikiText-2. This base is broadcast once at the start of
federation; thereafter only latents (ours) or the corresponding per-method payload
(baselines) are communicated.

\paragraph{Federated configuration.} We simulate $K$ clients ($K \in \{4,8,16,32\}$,
default $K=8$) over $R$ communication rounds with $E$ local epochs per round.
Client data is partitioned either \textbf{IID} (uniform random) or
\textbf{non-IID} via a Dirichlet split with concentration $\alpha \in \{0.5, 0.1\}$
(smaller $\alpha$ = more heterogeneous). Each round, clients start from the global
state, train locally, and upload their payload for averaging
(Algorithm~\ref{alg:latentfed}). The averaging identity is re-checked at every run.

\paragraph{Methods compared.} On a single bandwidth--accuracy axis we compare:
\begin{itemize}
  \item \textbf{Latent (ours):} delta low-rank mapping, default latent dimension
        $d=64$ per layer and rank $r=32$ on CIFAR-100 ($D = 1{,}280$ latents total),
        with a $d/r$ sweep for the Pareto curve. Trained with Adam
        (lr $10^{-2}$, weight decay $10^{-4}$, gradient clip $1.0$) on the latents only.
  \item \textbf{Full-weight FedAvg} \citep{mcmahan2017communication}: the standard
        baseline, communicating all trainable weights.
  \item \textbf{PowerSGD} \citep{vogels2019powersgd}: low-rank gradient compression
        with rank $\in \{2, 8\}$ and error feedback.
  \item \textbf{Top-$k$} \citep{aji2017sparse}: magnitude sparsification of the
        update at densities $\in \{1\%, 10\%\}$ with error feedback.
\end{itemize}
Except for the head-to-head study in \S\ref{ssec:res-h2h}, all baselines share
the same frozen pretrained base and frozen orthogonal head, so differences
reflect the communication scheme alone; the head-to-head study additionally
reports FedAvg with a trainable Kaiming head to isolate the head's
contribution.

\paragraph{Communication accounting.} We report \textbf{cumulative bytes per
client}, counting both upload and broadcast directions (\texttt{fp32} = $4$ bytes).
For top-$k$ we count transmitted values \emph{and} their integer indices; for
PowerSGD we count both transmitted factors. The per-round latent payload is
$D \cdot 4$ bytes ($\approx 5$\,KB for $D=1{,}280$); the per-round full-weight
payload is $\approx 45$\,MB.\footnote{We report bytes, not wall-clock time: the
experiments are a faithful federated simulation with exact byte accounting, but we
make no latency or system-throughput claims.}

\paragraph{Ablations and protocol.} The frozen-classifier ablation
(\S\ref{ssec:fc}) compares a trainable head, a frozen Kaiming-initialised head, and
our frozen orthogonal head, each over $3$ seeds. Unless stated otherwise, all
federated results are reported over $3$ seeds (mean $\pm$ standard deviation).
Pretraining and mapping both checkpoint per epoch and are resumable, and every run
logs the latent-averaging identity error as a correctness check.

\section{Results}
\label{sec:results}

We organise the results around the three claims of \S\ref{sec:intro}: massive
bandwidth reduction at parity accuracy (\S\ref{ssec:res-headline}--\ref{ssec:res-pareto}),
robustness under heterogeneity and client count (\S\ref{ssec:res-noniid}--\ref{ssec:res-k}),
and the supporting design choices (\S\ref{ssec:res-ablations}--\ref{ssec:res-scratch}).
All federated numbers are over $3$ seeds (mean $\pm$ std) unless noted. Additional
plots (fixed-byte budget, $K$-sweep, quantization Pareto, WikiText-2 language modelling)
and the full appendix appear in
Appendix~\ref{sup:extra-figs}--\ref{app:repro}.

\subsection{Federated fidelity at extreme compression}
\label{ssec:res-headline}

Table~\ref{tab:headline} reports the headline federated result on CIFAR-100
(ResNet-18+GN, $K=8$, IID). Communicating only the latent --- $1{,}280$ floats
($\approx 5$\,KB) per client per round, an $8718\times$ reduction relative to the
$11.2$M-parameter model --- our method reaches
$74.67 \pm 0.01\%$, within $\approx 0.5$\,pp of full-weight FedAvg
($75.16 \pm 0.15\%$) and statistically indistinguishable from the centrally
pretrained ceiling ($74.6\%$). The latent carries essentially all of the
federated signal at a per-round payload four orders of magnitude smaller than
the model.

\begin{table}[t]
  \centering
  \caption{Federated CIFAR-100 (ResNet-18+GN, $K=8$, IID, $3$ seeds). Our latent
  exchange matches full-weight FedAvg within seed noise at $8718\times$ fewer
  transmitted parameters per round.}
  \label{tab:headline}
  \begin{tabular}{lccc}
    \toprule
    Method & Test acc.\ (\%) & Payload/round/client & Compression \\
    \midrule
    Full-weight FedAvg & $75.16 \pm 0.15$ & $\approx 45$\,MB & $1\times$ \\
    \textbf{Latent (ours, $d{=}64$)} & $\mathbf{74.67 \pm 0.01}$ & $\approx 5$\,KB & $\mathbf{8718\times}$ \\
    Centralized base (ceiling) & $74.6$ & --- & --- \\
    \bottomrule
  \end{tabular}
\end{table}

\subsection{Head-to-head: ours vs textbook FedAvg}
\label{ssec:res-h2h}

\begin{figure}[t]
  \centering
  \includegraphics[width=0.7\linewidth]{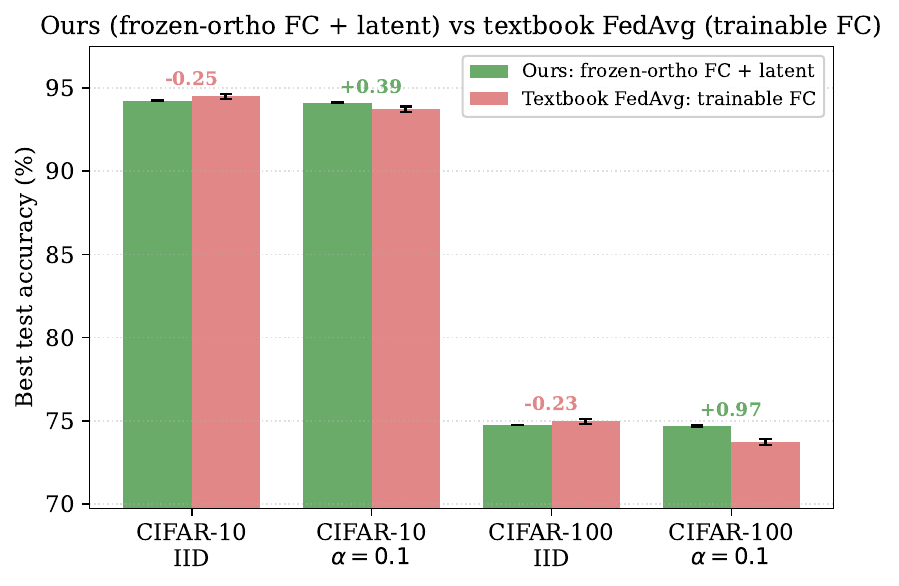}
  \caption{Our scheme (frozen-orthogonal head $+$ latent) vs textbook FedAvg
  (trainable head, full-weight). Parity under IID, a clear advantage under
  non-IID ($\alpha{=}0.1$), at a fraction of the bandwidth.}
  \label{fig:headtohead}
\end{figure}

Table~\ref{tab:headtohead} and Fig.~\ref{fig:headtohead} summarise the central
comparison across both datasets and partitions: our full scheme
(frozen orthogonal head $+$ latent) against textbook FedAvg (trainable Kaiming
head, full-weight averaging; the only experiment where the FedAvg baseline
uses a trainable Kaiming head --- elsewhere the head is matched to isolate
the communication channel). Under IID data our method is within seed noise
of FedAvg ($-0.25$\,pp on CIFAR-10, $-0.23$\,pp on CIFAR-100) at
$540$--$8700\times$ lower per-round bandwidth. Under strong heterogeneity
($\alpha{=}0.1$) the comparison \emph{reverses}: we are $+0.39$\,pp ahead on
CIFAR-10 and $+0.97$\,pp ahead on CIFAR-100.

% Auto-generated by make_figures.py:write_tables(). Do not edit by hand.
\begin{table}[t]
  \centering
  \caption{Ours (frozen orthogonal FC head + latent communication) vs.\ textbook FedAvg (trainable Kaiming FC, full-weight averaging). Best test accuracy (\%) (mean over 3 seeds); $\Delta=$ ours $-$ FedAvg (positive favours ours).}
  \label{tab:headtohead}
  \begin{tabular}{lccc}
    \toprule
    Dataset / partition & Latent (ours) & Textbook FedAvg & $\Delta$ \\
    \midrule
  CIFAR-10, IID & 94.23$\pm$0.02 & 94.47$\pm$0.14 & -0.25 \\
  CIFAR-10, $\alpha{=}0.1$ & 94.10$\pm$0.04 & 93.71$\pm$0.16 & \bfseries +0.39 \\
  CIFAR-100, IID & 74.74$\pm$0.01 & 74.97$\pm$0.15 & -0.23 \\
  CIFAR-100, $\alpha{=}0.1$ & 74.72$\pm$0.05 & 73.75$\pm$0.18 & \bfseries +0.97 \\
    \bottomrule
  \end{tabular}
\end{table}

\subsection{Bandwidth--accuracy Pareto}
\label{ssec:res-pareto}

\begin{figure}[t]
  \centering
  \begin{minipage}{0.49\linewidth}\centering
    \includegraphics[width=\linewidth]{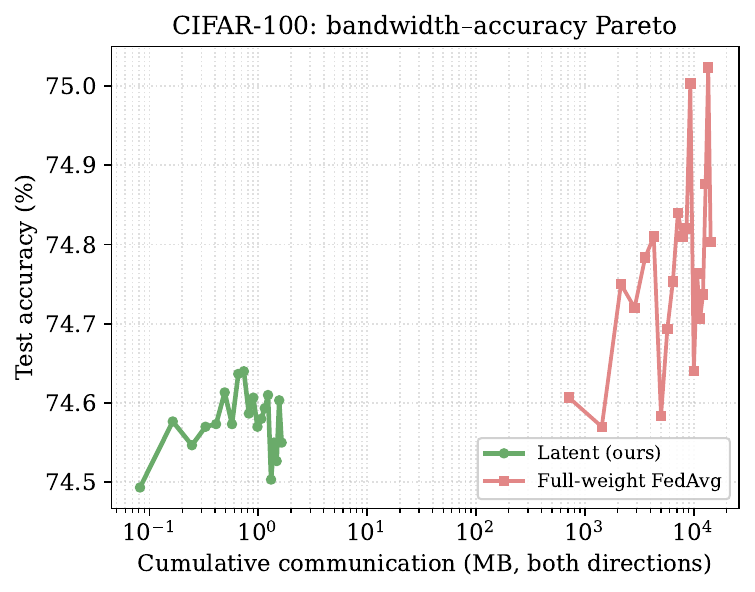}
  \end{minipage}\hfill
  \begin{minipage}{0.49\linewidth}\centering
    \includegraphics[width=\linewidth]{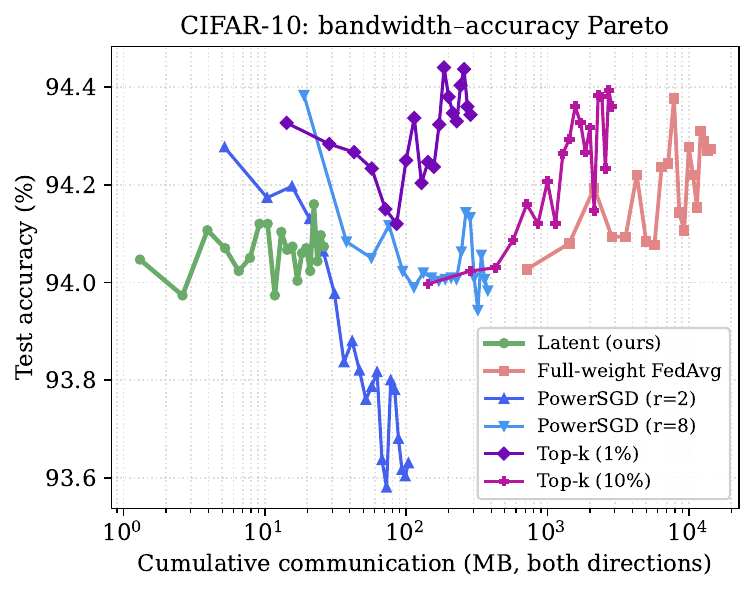}
  \end{minipage}
  \caption{Bandwidth--accuracy Pareto. Left: CIFAR-100, latent (ours) vs
  full-weight FedAvg. Right: CIFAR-10, vs full-weight, PowerSGD ($r\in\{2,8\}$) and
  top-$k$ ($\{1\%,10\%\}$). Our latent occupies the low-bandwidth frontier.}
  \label{fig:pareto}
\end{figure}

Figure~\ref{fig:pareto} plots test accuracy against cumulative communication.
On CIFAR-100 our latent points occupy the bottom-left of the plot --- the entire
trajectory lives below $\approx 2$\,MB cumulative, while full-weight FedAvg
requires $10^3$--$10^4$\,MB to reach comparable accuracy. The latent dimension
$d$ traces a controllable frontier (Table~\ref{tab:dsweep}): from $d=64$
($8718\times$, $74.67\%$) up to $d=1024$ ($545\times$, $74.66\%$), accuracy is
essentially flat, confirming that the correction a pretrained model needs is
genuinely small. On CIFAR-10 the same picture holds against the
stronger compression baselines: PowerSGD and top-$k$ reduce full-weight cost by
$40$--$140\times$, but our latent is a further $1$--$2$ orders of magnitude
cheaper at matched accuracy.

\begin{table}[t]
  \centering
  \caption{Latent-dimension sweep, federated CIFAR-100 ($K=8$, IID). Accuracy is
  flat across two orders of magnitude of per-round payload.}
  \label{tab:dsweep}
  \begin{tabular}{lcccc}
    \toprule
    Config & Latents $D$ & Payload/round & Compression & Test acc.\ (\%) \\
    \midrule
    Latent $d{=}64$   & $1{,}280$  & $\approx 5$\,KB   & $8718\times$ & $74.67 \pm 0.01$ \\
    Latent $d{=}256$  & $5{,}120$  & $\approx 20$\,KB  & $2180\times$ & $74.66$ \\
    Latent $d{=}1024$ & $20{,}480$ & $\approx 82$\,KB  & $545\times$  & $74.66$ \\
    Full-weight FedAvg & $11.2$M & $\approx 45$\,MB & $1\times$ & $75.16 \pm 0.15$ \\
    \bottomrule
  \end{tabular}
\end{table}

\subsection{Robustness to non-IID data}
\label{ssec:res-noniid}

\begin{figure}[t]
  \centering
  \includegraphics[width=0.55\linewidth]{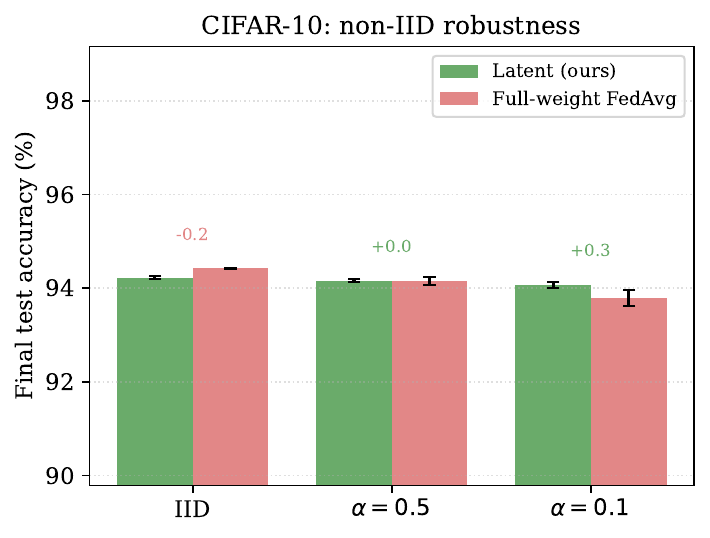}
  \caption{Non-IID robustness on CIFAR-10. Our latent matches full-weight FedAvg at
  $\alpha=0.5$ and exceeds it under strong heterogeneity ($\alpha=0.1$), at
  $\approx 545\times$ lower per-round bandwidth.}
  \label{fig:noniid}
\end{figure}

Figure~\ref{fig:noniid} compares our method with full-weight FedAvg on CIFAR-10
across IID and Dirichlet $\alpha \in \{0.5, 0.1\}$ splits. The gap to full-weight,
already small under IID ($-0.2$\,pp), vanishes at $\alpha=0.5$ and \emph{reverses}
under the strongest skew: at $\alpha=0.1$ our latent reaches $94.07 \pm 0.07\%$
versus full-weight $93.79 \pm 0.18\%$ ($+0.28$\,pp). Because the latent moves in
a low-dimensional shared subspace anchored at the pretrained base, client updates
are constrained and average more stably than full-weight updates --- exactly
the regime where full-weight FedAvg suffers most from client drift.

\subsection{Fixed-byte budget and client-count sensitivity}
\label{ssec:res-k}

Two further axes support the same picture (details in
Appendix~\ref{sup:extra-figs}):
under a fixed $200$\,MB per-client budget on CIFAR-10 ($\alpha{=}0.5$), our
method completes $\sim$$150$ latent rounds and reaches $94.21\%$ while
full-weight FedAvg can afford one round and attains $93.48\%$ ($+0.73$\,pp for
ours). Sweeping $K \in \{4,8,16,32\}$ on CIFAR-100, our latent accuracy is
essentially flat in $K$, whereas full-weight under $\alpha{=}0.5$ degrades to
$74.2\%$ at $K{=}4$ before recovering at large $K$ --- a further symptom of
drift that our low-dimensional channel avoids.

\subsection{Design ablations}
\label{ssec:res-ablations}

\paragraph{Frozen orthogonal classifier.} On CIFAR-100 (3 seeds), freezing a
Kaiming-initialised head is accuracy-neutral ($72.95\%$ vs trainable $72.99\%$),
but freezing an \emph{orthogonal} head improves accuracy to $73.53\%$
($+0.54$\,pp) while removing it from the payload. A fixed, well-conditioned set
of class prototypes gives the backbone a clean target geometry to align to at
zero communication cost.

\paragraph{Rank irrelevance under delta mode.} Consistent with \S\ref{ssec:delta},
the projection rank $r$ has little effect on accuracy in delta mode: varying
$r \in \{8, 32, 128, 256\}$ at $d{=}256$ on CIFAR-100 moves test accuracy by
$<0.2$\,pp. Rank therefore trades only generator memory and compute, not
accuracy --- justifying the small $r{=}32$ used throughout.

\subsection{Negative result: mapping networks do not scale as stand-alone training}
\label{ssec:res-scratch}

The original mapping-network proposal targets stand-alone training: a tiny
trainable latent, generated through a frozen random projection, replaces direct
weight optimisation. We find this does not extend beyond small architectures.
At ResNet-18 scale on CIFAR-100, training the latent from a randomly initialised
base collapses to $\approx 2.5\%$ test accuracy, against $74.7\%$ that direct
SGD reaches and that our delta-mode latent recovers. The collapse is robust:
it persists across latent dimensions $d\in\{1024, 4096, 16384\}$, projection
ranks $r\in\{8, 32, 128, 256\}$, and warm-start variants including
PCANet-initialised bases, short SGD warm-ups, and least-squares latent
initialisation from a converged teacher. The latent has ample capacity to encode
a \emph{correction} around a strong base, but not to drive a large network from
scratch.

This is the negative result that shapes our framing: rather than competing with
direct training, mapping networks are useful precisely as a low-bandwidth
\emph{communication channel} between models that have already been trained.

\subsection{Scaling to TinyImageNet}
\label{ssec:res-tin}

To test whether the trade-off holds at larger scale, we federate a
\textbf{ResNet-34+GN} backbone ($\approx 21.3$M parameters, $200$ classes) on
\textbf{TinyImageNet} with $d{=}1024$ ($K{=}8$, $R{=}20$, $2$ local epochs).
Under IID partitioning, the latent channel matches full-weight FedAvg within
$0.2$\,pp ($62.14 \pm 0.08\%$ vs $62.34 \pm 0.11\%$) while transmitting
$\approx 144$\,KB per client per round vs $\approx 81$\,MB
($\approx 577\times$ less). Under non-IID data ($\alpha{=}0.5$), the latent
path is \emph{more} accurate ($62.22 \pm 0.10\%$ vs $60.79 \pm 0.19\%$,
$+1.43$\,pp), echoing the CIFAR head-to-head trend at a larger scale
(Table~\ref{tab:tin}).

% Auto-generated by make_figures.py:write_tables(). Do not edit by hand.
\begin{table}[t]
  \centering
  \caption{TinyImageNet (ResNet-34+GN, $d{=}1024$, $K{=}8$, $R{=}20$): latent vs full-weight FedAvg. Best test accuracy (\%) (mean over 3 seeds); $\Delta=$ latent $-$ full-weight. Per-round payload: latent 144\,KB vs full-weight 81.2\,MB.}
  \label{tab:tin}
  \begin{tabular}{lccc}
    \toprule
    Partition & Latent (ours) & Full-weight FedAvg & $\Delta$ \\
    \midrule
  IID & 62.14$\pm$0.08 & 62.34$\pm$0.11 & -0.20 \\
  non-IID ($\alpha{=}0.5$) & 62.22$\pm$0.10 & 60.79$\pm$0.19 & \bfseries +1.42 \\
    \bottomrule
  \end{tabular}
\end{table}

\subsection{Integer-quantized latents}
\label{ssec:res-quant}

The latent is a short vector, so it composes directly with standard payload
compression. We apply uniform-affine integer quantization to the uploaded latent
and compare against quantizing the full-weight update at \emph{matched}
bit-widths (Table~\ref{tab:quant}; see Fig.~\ref{sup:fig:quant-bits} in the appendix for the accuracy-vs-precision plot). On CIFAR-100,
quantizing our latent to int8 or \emph{int4} leaves accuracy essentially
unchanged ($74.73 \pm 0.02\%$ at int4 vs $74.72 \pm 0.03\%$ at fp32, $3$ seeds),
shrinking the payload to $648$ bytes per client per round --- a $68{,}943\times$
reduction relative to the fp32 full-weight model. By contrast, textbook
full-weight FedAvg tolerates int8 but \emph{collapses} at int4 (to
$1.11 \pm 0.02\%$, i.e.\ chance), because $4$-bit quantization of $11.2$M
weights injects far more error than averaging can absorb. Quantization and our
latent channel are complementary and together push the per-round payload below
$1$\,KB at full accuracy.

% Auto-generated by make_figures.py:write_tables(). Do not edit by hand.
\begin{table}[t]
  \centering
  \caption{CIFAR-100 ($d{=}64$) matched-bit-width communication ladder. Payload is per client per round; best test accuracy (\%) (mean over 3 seeds). Our latent scheme is unaffected by 8/4-bit quantization, while textbook full-weight FedAvg collapses at int4 ($\dagger$).}
  \label{tab:quant}
  \begin{tabular}{llrcc}
    \toprule
    Method & Bits & Payload/client & Best acc.\ (\%) & $n$ \\
    \midrule
    Latent (ours) & fp32 & 5.0\,KB & 74.72$\pm$0.03 & 3 \\
    Latent (ours) & int8 & 1.3\,KB & 74.74$\pm$0.05 & 3 \\
    \bfseries Latent (ours) & int4 & 648.0\,B & 74.73$\pm$0.02 & 3 \\
    Full-weight FedAvg & fp32 & 42.6\,MB & 75.12$\pm$0.09 & 3 \\
    Full-weight FedAvg & int8 & 10.7\,MB & 74.80$\pm$0.17 & 3 \\
    Full-weight FedAvg & int4 & 5.3\,MB & 1.11$\pm$0.02\,$\dagger$ & 3 \\
    \bottomrule
  \end{tabular}
\end{table}

\section{Conclusion}
\label{sec:discussion}

\paragraph{Why it helps under heterogeneity.} Latent updates live in a
low-dimensional subspace anchored at the shared base, so client updates are
implicitly constrained and average more stably than full-weight updates. The
gap to full-weight FedAvg thus closes and then reverses as data become more
non-IID (\S\ref{ssec:res-noniid}); because each round is cheap, the method can
also synchronise often within a fixed byte budget, directly attacking drift
(\S\ref{ssec:res-k}). The empirical flatness of accuracy in both $d$
(Table~\ref{tab:dsweep}) and $r$ (\S\ref{ssec:res-ablations}) is consistent
with adapting a well-pretrained network being intrinsically low-dimensional
\citep{li2018intrinsic}: the latent encodes only a correction around a strong
base --- which is also why the from-scratch variant fails
(\S\ref{ssec:res-scratch}).

\paragraph{Composability.} The frozen orthogonal head, seed regeneration, and
latent are modular: any can be dropped without breaking the others, and the
latent itself composes with standard payload compression --- integer
quantization pushes the payload to $648$\,B/round at parity, where matched-bit
full-weight FedAvg collapses (\S\ref{ssec:res-quant}). The shared
centrally-pretrained base fits current federated practice, in that many
practical deployments adapt off public checkpoints rather than train from
scratch.

\paragraph{Limitations.} (i)~GroupNorm places \emph{absolute} accuracy below
BatchNorm SOTA; our claims are relative parity at massive compression, not
absolute state of the art. (ii)~The method is federated \emph{fine-tuning}: it
requires a shared centrally-pretrained base and a one-time broadcast of that
base and the seed. (iii)~Our evaluation is a federated simulation with exact
byte accounting; we make no wall-clock claims. A preliminary autoregressive
language-modelling extension is in Appendix~\ref{sup:wt2}.

\paragraph{Code.} The code will be released publicly upon acceptance of the
paper.

\bibliographystyle{splncs04}
\bibliography{refs}

@article{fedmd2019,
  title = {FedMD: Heterogeneous Federated Learning via Model Distillation},
  author = {Li, Daliang and Wang, Junpu},
  journal = {arXiv preprint arXiv:1910.03581},
  year = {2019},
  note = {NeurIPS 2019 Workshop on Federated Learning for Data Privacy and Confidentiality},
}

@inproceedings{dspodfl2025,
  title = {Decentralized Sporadic Federated Learning: A Unified Algorithmic Framework with Convergence Guarantees},
  author = {Zehtabi, Shahryar and Han, Dong-Jun and Parasnis, Rohit and Hosseinalipour, Seyyedali and Brinton, Christopher},
  booktitle = {International Conference on Learning Representations},
  year = {2025},
  note = {Spotlight},
}

@inproceedings{dpfl2025,
  title = {DPFL: Decentralized Personalized Federated Learning},
  author = {Kharrat, Salma and Canini, Marco and Horv{\'a}th, Samuel},
  booktitle = {Proceedings of The 28th International Conference on Artificial Intelligence and Statistics},
  series = {Proceedings of Machine Learning Research},
  volume = {258},
  pages = {5086--5094},
  year = {2025},
  publisher = {PMLR},
}

@inproceedings{ntkdfl2025,
  title = {NTK-DFL: Enhancing Decentralized Federated Learning in Heterogeneous Settings via Neural Tangent Kernel},
  author = {Thompson, Gabriel and Yue, Kai and Wong, Chau-Wai and Dai, Huaiyu},
  booktitle = {Proceedings of the 42nd International Conference on Machine Learning},
  series = {Proceedings of Machine Learning Research},
  volume = {267},
  pages = {59470--59491},
  year = {2025},
  publisher = {PMLR},
}

@inproceedings{fedspd2025,
  title = {FedSPD: A Soft-clustering Approach for Personalized Decentralized Federated Learning},
  author = {Lin, I-Cheng and Yagan, Osman and Joe-Wong, Carlee},
  booktitle = {Proceedings of the Forty-first Conference on Uncertainty in Artificial Intelligence},
  series = {Proceedings of Machine Learning Research},
  volume = {286},
  pages = {2618--2641},
  year = {2025},
  publisher = {PMLR},
}

@article{gflat2026,
  title = {Achieving Global Flatness in Decentralized Learning with Heterogeneous Data},
  author = {Choudhary, Sakshi and Aketi, Sai Aparna and Roy, Kaushik},
  journal = {Transactions on Machine Learning Research},
  year = {2026},
  note = {Accepted by TMLR}
}

@inproceedings{mcmahan2017communication,
  title = {Communication-Efficient Learning of Deep Networks from Decentralized Data},
  author = {McMahan, H. Brendan and Moore, Eider and Ramage, Daniel and Hampson, Seth and Ag{\"u}era y Arcas, Blaise},
  booktitle = {Proceedings of the 20th International Conference on Artificial Intelligence and Statistics (AISTATS)},
  series = {Proceedings of Machine Learning Research},
  volume = {54},
  pages = {1273--1282},
  year = {2017},
  publisher = {PMLR},
}

@article{douillard2023diloco,
  title = {{DiLoCo}: Distributed Low-Communication Training of Language Models},
  author = {Douillard, Arthur and Feng, Qixuan and Rusu, Andrei A. and Chhaparia, Rachita and Donchev, Yani and Kuncoro, Adhiguna and Ranzato, Marc'Aurelio and Szlam, Arthur and Shen, Jiajun},
  journal = {arXiv preprint arXiv:2311.08105},
  year = {2023},
}

@inproceedings{stich2018local,
  title = {Local {SGD} Converges Fast and Communicates Little},
  author = {Stich, Sebastian U.},
  booktitle = {International Conference on Learning Representations},
  year = {2019},
}

@inproceedings{khaled2020tighter,
  title = {Tighter Theory for Local {SGD} on Identical and Heterogeneous Data},
  author = {Khaled, Ahmed and Mishchenko, Konstantin and Richt{\'a}rik, Peter},
  booktitle = {Proceedings of the 23rd International Conference on Artificial Intelligence and Statistics (AISTATS)},
  series = {Proceedings of Machine Learning Research},
  volume = {108},
  pages = {4519--4529},
  year = {2020},
  publisher = {PMLR},
}

@article{lin2020ensemble,
  title={Ensemble Distillation for Robust Model Fusion in Federated Learning},
  author={Lin, Tao and Kong, Lingjing and Stich, Sebastian U. and Jaggi, Martin},
  journal={arXiv preprint arXiv:2006.07242},
  year={2020}
}

@article{kairouz2019advances,
  title={Advances and Open Problems in Federated Learning},
  author={Kairouz, Peter and McMahan, H. Brendan and Avent, Brendan and Bellet, Aur{\'e}lien and Bennis, Mehdi and Bhagoji, Arjun Nitin and Bonawitz, Keith and Charles, Zachary and Cormode, Graham and Cummings, Rachel and others},
  journal={arXiv preprint arXiv:1912.04977},
  year={2019}
}

@article{li2018federated,
  title={Federated Optimization in Heterogeneous Networks},
  author={Li, Tian and Sahu, Anit Kumar and Zaheer, Manzil and Sanjabi, Maziar and Talwalkar, Ameet and Smith, Virginia},
  journal={arXiv preprint arXiv:1812.06127},
  year={2018}
}

@article{karimireddy2019scaffold,
  title={SCAFFOLD: Stochastic Controlled Averaging for On-Device Federated Learning},
  author={Karimireddy, Sai Praneeth and Kale, Satyen and Mohri, Mehryar and Reddi, Sashank J. and Stich, Sebastian U. and Suresh, Ananda Theertha},
  journal={arXiv preprint arXiv:1910.06378},
  year={2019}
}

@article{reddi2020adaptive,
  title={Adaptive Federated Optimization},
  author={Reddi, Sashank and Charles, Zachary and Zaheer, Manzil and Garrett, Zachary and Rush, Keith and Kone{\v{c}}n{\'y}, Jakub and Kumar, Sanjiv and McMahan, H. Brendan},
  journal={arXiv preprint arXiv:2003.00295},
  year={2020}
}

@inproceedings{wang2020federated,
  title={Federated Learning with Matched Averaging},
  author={Wang, Hongyi and Yurochkin, Mikhail and Sun, Yuekai and Papailiopoulos, Dimitris and Khazaeni, Yasaman},
  booktitle={International Conference on Learning Representations},
  year={2020}
}

@inproceedings{vogels2019powersgd,
  title     = {{PowerSGD}: Practical Low-Rank Gradient Compression for Distributed Optimization},
  author    = {Vogels, Thijs and Karimireddy, Sai Praneeth and Jaggi, Martin},
  booktitle = {Advances in Neural Information Processing Systems 32 (NeurIPS)},
  year      = {2019},
}

@inproceedings{aji2017sparse,
  title     = {Sparse Communication for Distributed Gradient Descent},
  author    = {Aji, Alham Fikri and Heafield, Kenneth},
  booktitle = {Proceedings of the 2017 Conference on Empirical Methods in Natural Language Processing (EMNLP)},
  pages     = {440--445},
  year      = {2017},
}

@inproceedings{alistarh2017qsgd,
  title     = {{QSGD}: Communication-Efficient {SGD} via Gradient Quantization and Encoding},
  author    = {Alistarh, Dan and Grubic, Demjan and Li, Jerry and Tomioka, Ryota and Vojnovic, Milan},
  booktitle = {Advances in Neural Information Processing Systems 30 (NeurIPS)},
  year      = {2017},
}

@inproceedings{bernstein2018signsgd,
  title     = {{signSGD}: Compressed Optimisation for Non-Convex Problems},
  author    = {Bernstein, Jeremy and Wang, Yu-Xiang and Azizzadenesheli, Kamyar and Anandkumar, Anima},
  booktitle = {Proceedings of the 35th International Conference on Machine Learning (ICML)},
  year      = {2018},
}

@inproceedings{lin2018deep,
  title     = {Deep Gradient Compression: Reducing the Communication Bandwidth for Distributed Training},
  author    = {Lin, Yujun and Han, Song and Mao, Huizi and Wang, Yu and Dally, William J.},
  booktitle = {International Conference on Learning Representations},
  year      = {2018},
}

@inproceedings{karimireddy2019error,
  title     = {Error Feedback Fixes {SignSGD} and other Gradient Compression Schemes},
  author    = {Karimireddy, Sai Praneeth and Rebjock, Quentin and Stich, Sebastian U. and Jaggi, Martin},
  booktitle = {Proceedings of the 36th International Conference on Machine Learning (ICML)},
  year      = {2019},
}

@inproceedings{wu2018group,
  title     = {Group Normalization},
  author    = {Wu, Yuxin and He, Kaiming},
  booktitle = {Proceedings of the European Conference on Computer Vision (ECCV)},
  year      = {2018},
}

@inproceedings{hsieh2020noniid,
  title     = {The Non-{IID} Data Quagmire of Decentralized Machine Learning},
  author    = {Hsieh, Kevin and Phanishayee, Amar and Mutlu, Onur and Gibbons, Phillip B.},
  booktitle = {Proceedings of the 37th International Conference on Machine Learning (ICML)},
  year      = {2020},
}

@inproceedings{li2018intrinsic,
  title     = {Measuring the Intrinsic Dimension of Objective Landscapes},
  author    = {Li, Chunyuan and Farkhoor, Heerad and Liu, Rosanne and Yosinski, Jason},
  booktitle = {International Conference on Learning Representations},
  year      = {2018},
}

@inproceedings{hu2021lora,
  title     = {{LoRA}: Low-Rank Adaptation of Large Language Models},
  author    = {Hu, Edward J. and Shen, Yelong and Wallis, Phillip and Allen-Zhu, Zeyuan and Li, Yuanzhi and Wang, Shean and Wang, Lu and Chen, Weizhu},
  booktitle = {International Conference on Learning Representations},
  year      = {2022},
}

@article{mappingnetworks2026,
  title   = {Mapping Networks: Generating Network Weights from Low-Dimensional Latents},
  author  = {Sen, Lord and Mukherjee, Shyamapada},
  journal = {arXiv preprint arXiv:2602.19134},
  year    = {2026},
}

@inproceedings{hoffer2018fix,
  title     = {Fix Your Classifier: The Marginal Value of Training the Last Weight Layer},
  author    = {Hoffer, Elad and Hubara, Itay and Soudry, Daniel},
  booktitle = {International Conference on Learning Representations},
  year      = {2018},
}

@article{pernici2021regular,
  title   = {Regular Polytope Networks},
  author  = {Pernici, Federico and Bruni, Matteo and Baecchi, Claudio and Del Bimbo, Alberto},
  journal = {IEEE Transactions on Neural Networks and Learning Systems},
  year    = {2021},
}

@inproceedings{oh2022fedbabu,
  title     = {{FedBABU}: Toward Enhanced Representation for Federated Image Classification},
  author    = {Oh, Jaehoon and Kim, Sangmook and Yun, Se-Young},
  booktitle = {International Conference on Learning Representations},
  year      = {2022},
}

@article{papyan2020prevalence,
  title   = {Prevalence of Neural Collapse during the Terminal Phase of Deep Learning Training},
  author  = {Papyan, Vardan and Han, X. Y. and Donoho, David L.},
  journal = {Proceedings of the National Academy of Sciences},
  volume  = {117},
  number  = {40},
  pages   = {24652--24663},
  year    = {2020},
}

@inproceedings{li2021fedbn,
  title     = {{FedBN}: Federated Learning on Non-{IID} Features via Local Batch Normalization},
  author    = {Li, Xiaoxiao and Jiang, Meirui and Zhang, Xiaofei and Kamp, Michael and Dou, Qi},
  booktitle = {International Conference on Learning Representations},
  year      = {2021},
}

\appendix
% ============================================================================
% Appendix / supplementary material for the arXiv version of FLITE.
% ---------------------------------------------------------------------------
% Cross-referenced from the main paper. Includes: fixed-byte-budget figure,
% K-sweep figure, ortho-FC ablation figure, quantization Pareto figure, the
% WikiText-2 language modelling subsection, and the full hyperparameter /
% memory / reproducibility appendix.
% ============================================================================

\section{Additional experimental figures}
\label{sup:extra-figs}

This appendix collects extra figures referenced by the main results
(\S\ref{sec:results}) but not shown there for length reasons.

\subsection{Fixed-byte budget}
\label{sup:sec:budget}

\begin{figure}[h]
  \centering
  \includegraphics[width=0.55\linewidth]{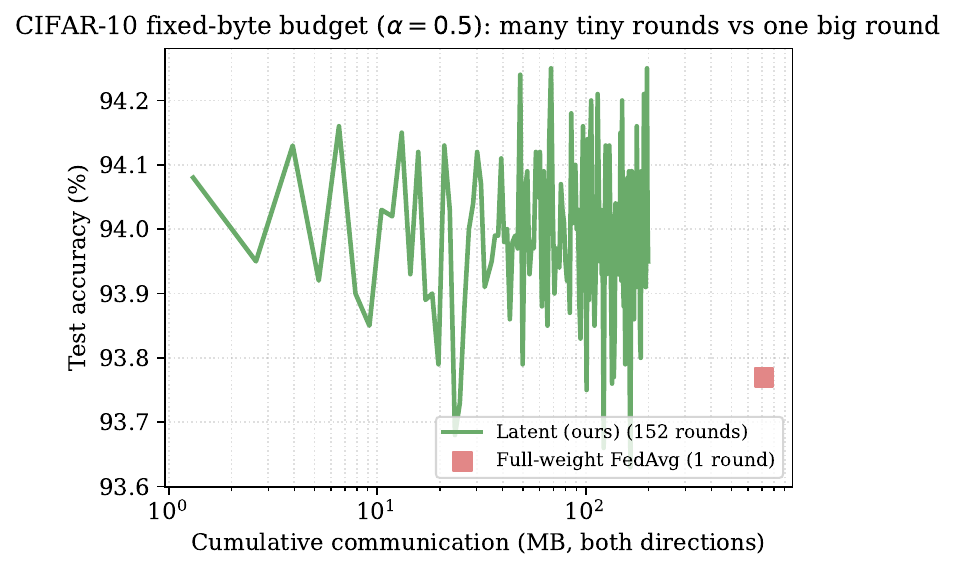}
  \caption{Fixed $200$\,MB/client budget (CIFAR-10, $\alpha{=}0.5$): many cheap
  latent rounds vs one expensive full-weight round.}
  \label{sup:fig:budget}
\end{figure}

Under a fixed $200$\,MB per-client budget on CIFAR-10 ($\alpha{=}0.5$), our
latent completes $\sim$$150$ rounds and reaches $94.21\%$ while full-weight
FedAvg can afford one round and attains $93.48\%$ ($+0.73$\,pp for ours,
Fig.~\ref{sup:fig:budget}). This is the fixed-budget view of
\S\ref{ssec:res-k}.

\subsection{Client-count sweep}
\label{sup:sec:ksweep}

\begin{figure}[h]
  \centering
  \includegraphics[width=0.55\linewidth]{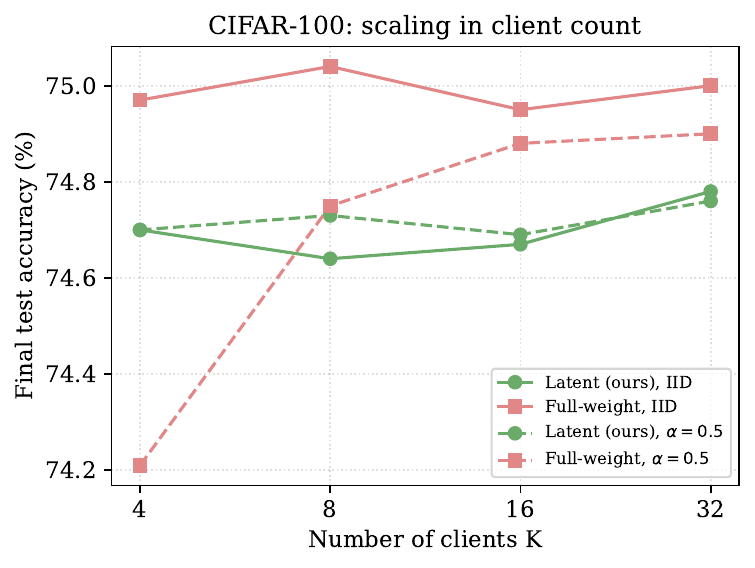}
  \caption{Scaling in client count $K$ (CIFAR-100). Latent accuracy is stable
  in $K$; full-weight is more sensitive at small $K$.}
  \label{sup:fig:ksweep}
\end{figure}

Figure~\ref{sup:fig:ksweep} sweeps $K \in \{4,8,16,32\}$ on CIFAR-100. Latent
accuracy is essentially flat in $K$ (and the per-round compression ratio is
$K$-invariant by construction). Full-weight FedAvg under $\alpha{=}0.5$
degrades at small $K$ (down to $74.2\%$ at $K{=}4$) and only recovers as $K$
grows, while our method is stable across the range. This complements the
client-count summary in \S\ref{ssec:res-k}.

\subsection{Frozen orthogonal head figure}
\label{sup:sec:orthofc}

\begin{figure}[h]
  \centering
  \includegraphics[width=0.5\linewidth]{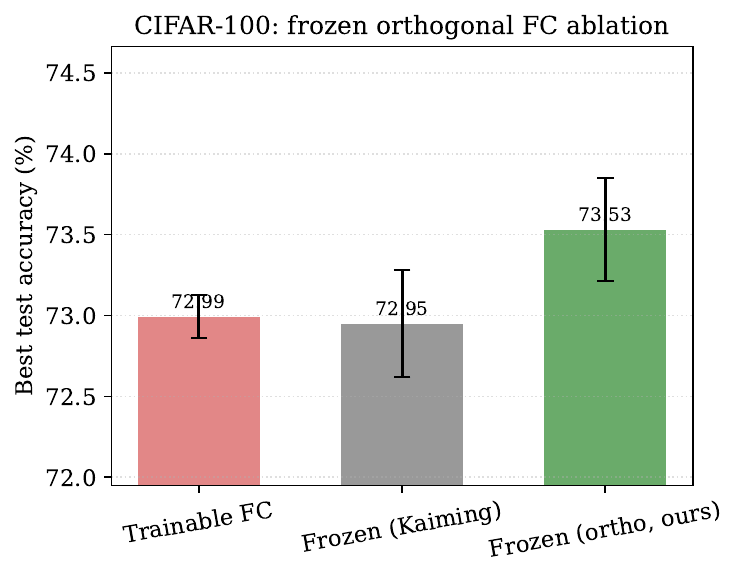}
  \caption{Classifier-head ablation on CIFAR-100 (frozen orthogonal vs.\
  trainable vs.\ frozen Kaiming).}
  \label{sup:fig:orthofc}
\end{figure}

Figure~\ref{sup:fig:orthofc} visualises the classifier-head ablation cited in
\S\ref{ssec:res-ablations}. The frozen orthogonal head improves accuracy over
both a trainable and a frozen Kaiming head, at zero communication cost.

\subsection{Quantization precision plot}
\label{sup:sec:quant-bits}

\begin{figure}[h]
  \centering
  \includegraphics[width=0.55\linewidth]{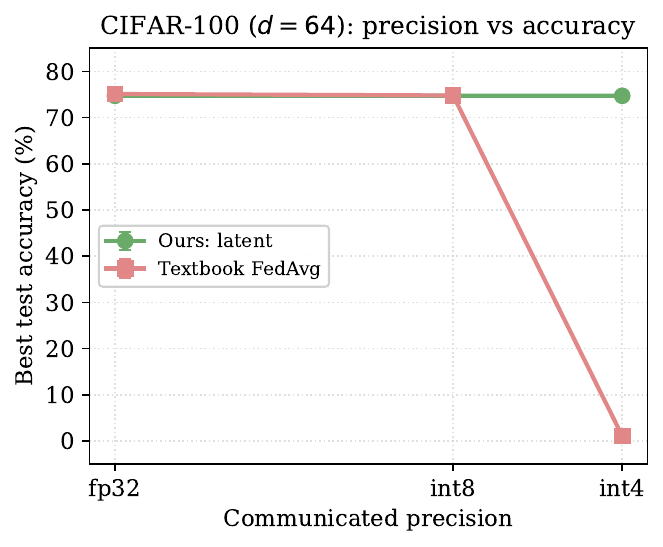}
  \caption{Latent quantization (CIFAR-100, $d{=}64$): accuracy vs communicated
  precision.}
  \label{sup:fig:quant-bits}
\end{figure}

Figure~\ref{sup:fig:quant-bits} shows accuracy vs.\ communicated precision on
CIFAR-100 ($d{=}64$), the plot complement of Table~\ref{tab:quant} in
\S\ref{ssec:res-quant}. Our latent is flat across fp32/int8/int4; full-weight
FedAvg collapses at int4.

\subsection{Quantization--bandwidth Pareto}
\label{sup:sec:quant-pareto}

\begin{figure}[h]
  \centering
  \includegraphics[width=0.55\linewidth]{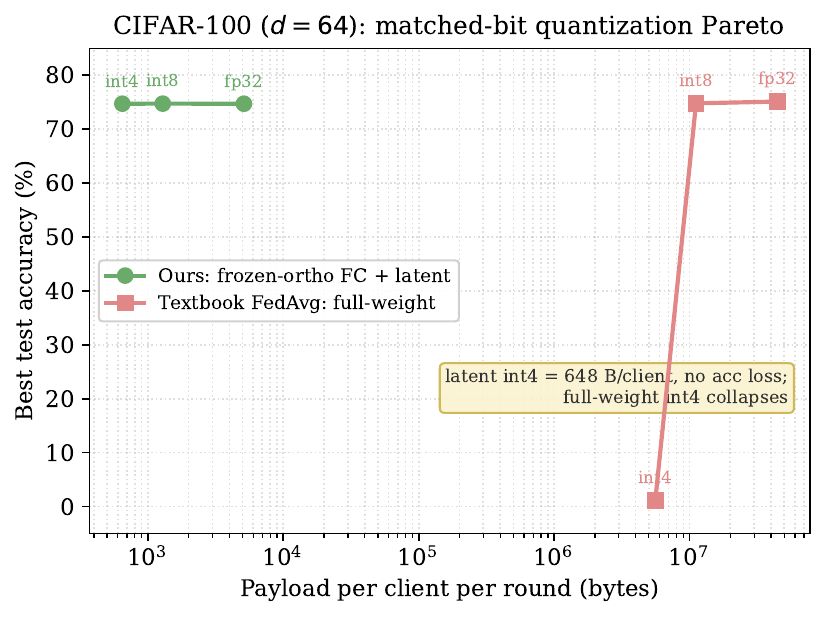}
  \caption{Latent-quantization bandwidth--accuracy points: int4 latent reaches
  $648$\,B/round at parity accuracy.}
  \label{sup:fig:quant-pareto}
\end{figure}

Combining the latent channel with integer quantization pushes the per-round
payload below $1$\,KB at parity accuracy (Fig.~\ref{sup:fig:quant-pareto}),
extending the Pareto discussion of \S\ref{ssec:res-pareto}.

% ============================================================================
\section{Language modelling on WikiText-2}
\label{sup:wt2}

\begin{figure}[h]
  \centering
  \includegraphics[width=0.62\linewidth]{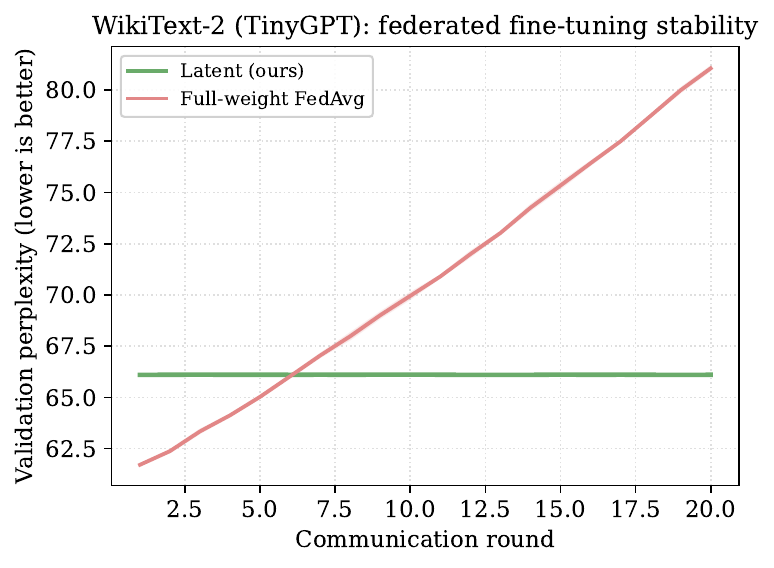}
  \caption{WikiText-2 (TinyGPT): validation perplexity vs.\ communication round.
  Shaded bands are $\pm 1$ std over $3$ seeds. Latent federation is stable at
  the pretrained base; full-weight FedAvg improves briefly then drifts.}
  \label{sup:fig:wt2}
\end{figure}

Beyond vision, we apply \methodname{} to federated language-model fine-tuning
on WikiText-2 with the TinyGPT backbone ($d{=}1024$, $K{=}8$, $R{=}20$,
$1$ local epoch, IID split; $3$ seeds). Table~\ref{tab:wt2} and
Fig.~\ref{sup:fig:wt2} report validation perplexity (lower is better).

The headline result here is \textbf{communication}: our latent channel transmits
$\approx 144$\,KB per client per round versus $\approx 44$\,MB for full-weight
FedAvg --- a $\approx 313\times$ reduction --- while holding validation
perplexity essentially at the shared pretrained base ($66.09$ throughout
federation). By contrast, full-weight FedAvg can briefly improve perplexity in
early rounds ($61.71 \pm 0.03$ best) but then \emph{diverges} as heterogeneous
local updates are averaged ($81.06 \pm 0.09$ final). We therefore treat
WikiText-2 primarily as evidence that the kilobyte-scale payload transfers to
autoregressive LMs; accuracy parity on this task is left to future tuning of
local steps and learning rates.

% Auto-generated by make_figures.py:write_tables(). Do not edit by hand.
\begin{table}[t]
  \centering
  \caption{WikiText-2 (TinyGPT, $d{=}1024$, $K{=}8$, $R{=}20$, IID). Per-round payload and validation perplexity (mean over 3 seeds). Latent payload is $313\times$ smaller. Lower perplexity is better.}
  \label{tab:wt2}
  \begin{tabular}{lrrr}
    \toprule
    Method & Payload/client & Best val PPL & Final val PPL \\
    \midrule
  Latent (ours) & 144\,KB & 66.09$\pm$0.00 & 66.10$\pm$0.00 \\
  Full-weight FedAvg & 44.0\,MB & 61.71$\pm$0.03 & 81.06$\pm$0.09 \\
    \bottomrule
  \end{tabular}
\end{table}

% ============================================================================
\section{Hyperparameters and training recipes}
\label{app:hparams}

\paragraph{Centralized pretraining (shared base).} SGD with momentum $0.9$,
initial learning rate $0.05$, $3$-epoch linear warmup then cosine decay to zero,
weight decay $5\times10^{-4}$, batch size $128$, $100$ epochs, standard
crop+flip augmentation. GroupNorm with $32$ groups. The classifier head is
orthogonally initialised and frozen from the first step. Per-epoch checkpointing
enables resume.

\paragraph{Latent (mapping) training.} Adam, learning rate $10^{-2}$, weight decay
$10^{-4}$, gradient clipping at $1.0$, latents initialised at zero (so the model
starts exactly at the pretrained base). Default $d=64$, $r=32$ on CIFAR-100.
Backbone, projection factors $U,V$, and the classifier head are all frozen; only
the per-layer latents $z_l$ are trained.

\paragraph{Federated configuration.} $K$ clients ($K\in\{4,8,16,32\}$, default
$8$), $E$ local epochs per round, IID or Dirichlet($\alpha$) partition with
$\alpha\in\{0.5,0.1\}$. Server averaging is the unweighted mean of client latents
(ours) or client payloads (baselines). Baselines: full-weight FedAvg; PowerSGD
rank $\in\{2,8\}$ with error feedback; top-$k$ density $\in\{1\%,10\%\}$ with error
feedback. Except for the head-to-head study in the main paper, all methods
share the same frozen pretrained base and the same frozen orthogonal head.

\section{Generator memory analysis}
\label{app:memory}

\begin{figure}[h]
  \centering
  \includegraphics[width=0.6\linewidth]{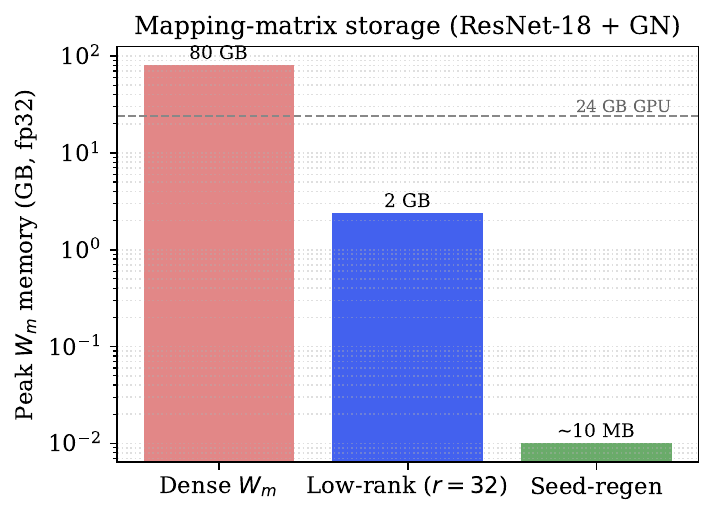}
  \caption{Generator storage for ResNet-18: dense ($\approx 80$\,GB) vs low-rank
  ($\approx 2$\,GB) vs seed-regenerated ($\approx 10$\,MB working memory).}
  \label{fig:memory}
\end{figure}

The original mapping-network generator stores a dense projection $W_m$ of shape
$P_l\times d_l$ for each mapped layer. Aggregated over ResNet-18
($\sum_l P_l \approx 11.2$M), even a modest per-layer $d_l$ makes the dense
projections require on the order of $80$\,GB in \texttt{fp32}, exceeding a single
accelerator. Our low-rank factorisation $W_m = U V^{\top}$ stores
$r\,(P_l + d_l)$ values per layer instead of $P_l\,d_l$; at $r=32$ this is
$\approx 2$\,GB. Seed regeneration stores neither factor --- a single $64$-bit
seed deterministically regenerates $U_l, V_l$ (Gaussian draw + QR) identically on
every client --- so only $\approx 10$\,MB of peak working memory is needed during
regeneration. Figure~\ref{fig:memory} compares these storage options. The exact
averaging identity is preserved in all variants because $U_l, V_l$ are identical
across clients (same seed $\Rightarrow$ same factors).

\section{From-scratch failure at scale}
\label{app:scratch}

Training the latent from a randomly (Kaiming) initialised base, with no pretrained
offset, collapses at ResNet-18 scale: on CIFAR-100 it reaches only
$\approx 2.5\%$ test accuracy, against $74.7\%$ in delta mode under otherwise
identical settings. The latent has the capacity to encode a correction around a
strong base, but not to drive a large network from scratch; this motivates the
delta-on-pretrained framing and our positioning of the method as low-bandwidth
federated \emph{fine-tuning}.

\section{Rank sweep (rank irrelevance under delta mode)}
\label{app:rank}

At fixed $d=256$ on CIFAR-100, varying the projection rank
$r\in\{8,32,128,256\}$ changes test accuracy by $<0.2$\,pp. The rank therefore
trades only generator memory and regeneration compute, not accuracy, consistent
with the delta a well-pretrained model needs being itself low-dimensional. We use
$r=32$ throughout.

\section{Reproducibility}
\label{app:repro}

Every run logs the latent-averaging identity error (maximum discrepancy between
averaging latents and averaging the generated weights), which is at the level of
floating-point round-off ($\le 6\times10^{-8}$) in all experiments. Pretraining and
mapping checkpoint per epoch and are resumable. Communication is accounted in bytes
(both directions, \texttt{fp32}), counting transmitted values and indices for
top-$k$ and both factors for PowerSGD. Code and configuration files will be released.

\end{document}